\documentclass{INTERSPEECH2023}
\usepackage{booktabs}
\usepackage{multirow}
\usepackage{rotating}

\interspeechcameraready

\title{Pre-Finetuning for Few-Shot Emotional Speech Recognition}
\name{Maximillian Chen, Zhou Yu}
\address{
  Columbia University
  }
\email{maxchen@cs.columbia.edu, zy2461@columbia.edu}

\begin{document}

\maketitle
 
\begin{abstract}
Speech models have long been known to overfit individual speakers for many classification tasks. This leads to poor generalization in settings where the speakers are out-of-domain or out-of-distribution, as is common in production environments. We view speaker adaptation as a few-shot learning problem and propose investigating transfer learning approaches inspired by recent success with pre-trained models in natural language tasks. We propose pre-finetuning speech models on difficult tasks to distill knowledge into few-shot downstream classification objectives. We pre-finetune Wav2Vec2.0 on every permutation of four multiclass emotional speech recognition corpora and evaluate our pre-finetuned models through 33,600 few-shot fine-tuning trials on the Emotional Speech Dataset. 
\end{abstract}

\section{Introduction}
Speech models tend to generalize poorly to out-of-distribution speakers due to a phenomenon called speaker overfitting~\cite{jung2018avoiding,pironkov2016speaker,wang2020spoken}. Speaker overfitting can be problematic when deploying systems in production environments. There, speakers typically do not exist in training corpora, and thus it takes time to amass sufficient amounts of training data. This motivates systems which can adapt to individual speakers ``on-the-fly'' with little data.

To this end, out-of-domain speaker adaptation can be viewed as a few-shot learning problem~\cite{wang2020spoken}. In low resource data settings, learning can be difficult, particularly with the regularization used in typical speaker-invariant learning approaches~\cite{wang2020spoken,zhao2019speech}. However, in recent years, many approaches to few-shot learning have found success with transfer learning, leveraging pre-trained models which have learned representations from multiple corpora~\cite{aghajanyan-etal-2021-muppet, gao2021making,brown2020language}. But, these pretraining corpora are not necessarily relevant to target downstream tasks. This has motivated recent work to propose an additional step between pre-training and downstream fine-tuning called pre-finetuning~\cite{aghajanyan-etal-2021-muppet}. While most work has examined pre-finetuning with multi-task learning, \cite{padmakumar-etal-2022-exploring} found that large-scale multi-task pre-finetuning performance can nearly be matched by only using corpora from tasks that match the type of the eventual downstream task. 

\begin{figure}
    \centering
    \scalebox{0.75}
    {\includegraphics[width=\linewidth]{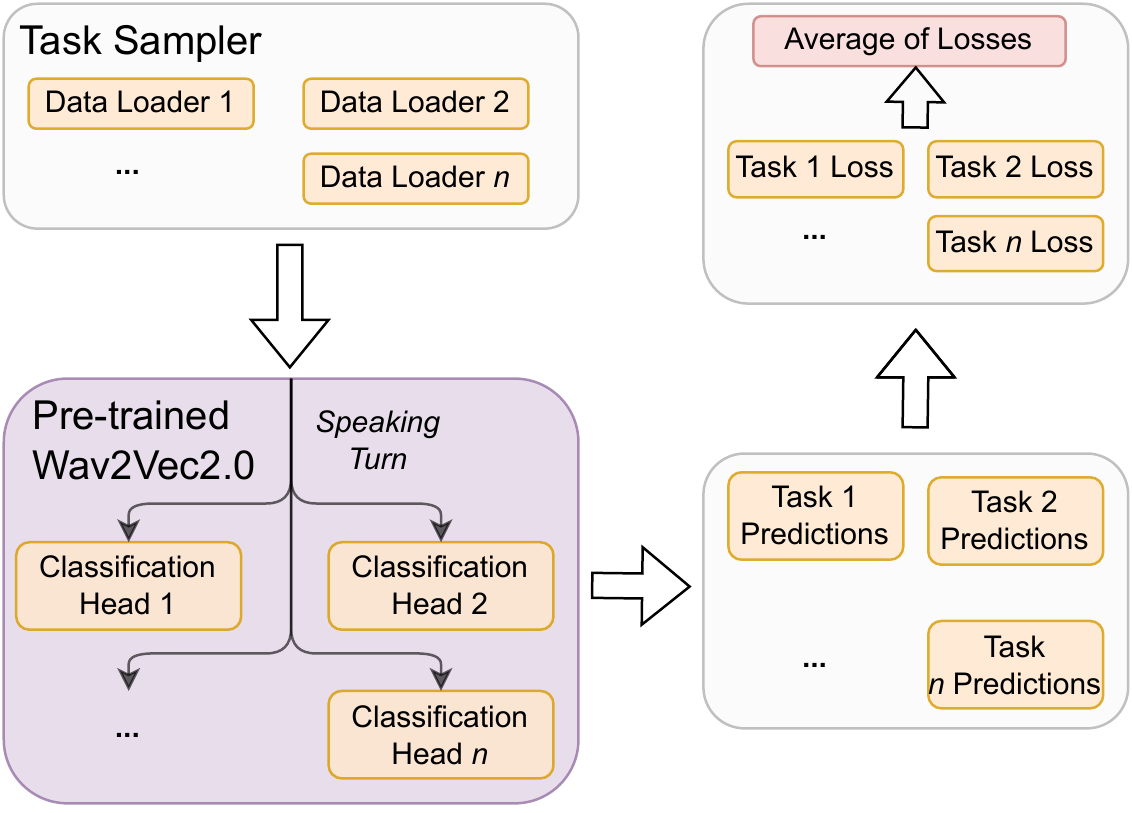}
    }
    \caption{Workflow of pre-finetuning an emotion recognition model. Wav2Vec2.0 is initialized with a separate linear classification head for each pre-finetuning dataset in order to ensure the correct output space. Pre-finetuning tasks are continuously randomly sampled, and each instance is mapped to the corresponding classification head. Each task's loss is computed separately and averaged during validation.}
    \label{fig:validation_epoch_workflow}
\end{figure}

Overall, pre-finetuning has been studied extensively and successfully in natural language processing tasks, but our study is the first to attempt pre-finetuning for any speech or audio processing task. We reason this is due to the comparatively higher number of tasks and datasets with accessible licensing for natural language. Additionally, many language tasks are easily cross-compatible during transfer learning (e.g. every pre-training task for T5 is trained using textual inputs and outputs without needing to adapt the  model or loss function~\cite{raffel2020exploring}). Here, we consider the few-shot\footnote{We use as few as two downstream training examples.} emotional speech recognition task.
We pre-finetune a separate model using every member of the power set of four large emotional speech corpora 
according to the workflow in Figure~\ref{fig:validation_epoch_workflow}.\footnote{Code at https://github.com/maxlchen/Speech-PreFinetuning} We evaluate our pre-finetuned models across 33,600 controlled few-shot classification experiments. We contribute ablations and analyses into how different experimental conditions affect pre-finetuning efficacy.
\section{Related Work}
Pre-trained models have been a successful case study of transfer learning for low resource tasks. When scaled to billions of parameters, pre-trained models have been able to generalize pre-trained knowledge to downstream tasks through few-shot in-context learning, taking the form of both downstream task models (e.g.~\cite{brown2020language,liu2023pre,min2022rethinking,wei2022emergent}) and models for synthetic data generation (e.g.~\cite{chen2022weakly,liu22low, meng2022generating,chen2023places}) in natural language tasks. However, capabilities such as in-context learning typically only appear through this tradeoff with model size, as they only emerge when using sufficiently large models. To date, pre-trained speech models have rarely been scaled to the same extent. The largest model is BigSSL with eight billion parameters~\cite{zhang2022bigssl}, whereas recent work on in-context learning for low-resource language data augmentation has used models with a \textit{minimum} size of six billion~\cite{chen2022weakly,sahu2022data,chen2023places}, and often reaching model sizes as large as 175 billion parameters~\cite{sahu2022data}. Smaller pre-trained models lack the same expansive pre-trained representations found in large models, making it impractical to perform in-context learning or generate high-quality synthetic data. However, due to their smaller size, it is more feasible to perform fine-tuning to directly transfer knowledge to downstream tasks.

But, fine-tuning still is impractical when there is not a sufficient amount of data. Recent work proposed an additional learning step between pre-training and fine-tuning to better facilitate knowledge distillation: pre-finetuning~\cite{aghajanyan-etal-2021-muppet}. They pre-finetuned the base and large variants of BART~\cite{lewis-etal-2020-bart} and RoBERTa~\cite{liu2019roberta} using large-scale multitask corpora. Follow-up work found that it was more efficient to do single-task pre-finetuning than to have a diverse set of corpora~\cite{padmakumar-etal-2022-exploring}. With the right corpora selection, pre-finetuning has potential to help with few-shot learning problems~\cite{ma2022label} such as speaker adaptation, but to date it has yet to be explored in any speech processing task. 

Traditional approaches to speaker adaptation focus on invariance using adversarial learning or regularization. \cite{meng2018speaker} proposed a network to jointly learn a speaker classifier and senone discriminator through adversarial multitask learning. Several works have investigated using KL divergence-based regularization (e.g. \cite{kim2017regularized,liu2016investigations}). Other studies used gradient reversal layers to regularize learning gains from individual speakers \cite{yin2020speaker}. But, these approaches often still require large amounts of training data and are not as applicable to low resource settings \cite{meng2018speaker,zhao2019speech}. 

The overall success with using pre-trained models in few-shot learning and increasing popularity of pre-trained speech models such as Wav2Vec2.0~\cite{baevski2020wav2vec} and HuBERT~\cite{hsu2021hubert} naturally motivates the exploration of pre-finetuning for few-shot speaker adaptation. The most similar lines of work examine multi-task learning for speech processing~\cite{chen2015multitask,cai2021speech,lirias3891890}, which also involves learning representations from multiple data sources. However, the key difference compared to our setting is that classic multitask learning involves training a model to learn a representation shared between a target downstream task and any auxiliary tasks \textit{simultaneously}~\cite{pironkov2016multi}. This requires sufficient downstream data. Pre-finetuning is a form of multitask learning which instead takes place during an intermediate step dedicated to learning an auxiliary task representation, which in turn can be used as a close initialization for a low-resource downstream task. Our work is the first to examine pre-finetuning speech models. We ground our study in the context of few-shot speaker adaptation for emotional speech recognition, and draw upon findings from \cite{padmakumar-etal-2022-exploring} in our selection of pre-finetuning corpora.

\section{Methodology}
\subsection{Corpora Selection}
\label{corpora_selection}
In this study, we focus on adapting speakers for emotion recognition as our downstream task. As such, we chose four large, diverse pre-finetuning corpora that each fall within the category of emotional speech recognition. MSP-IMPROV contains 8,438 improvised speaking turns from four emotions (happy, sadness, anger, neutral) \cite{busso2016msp}. MSP-PODCAST contains 100 hours of speech from 62,140 from speaking turns collected from podcast recordings~\cite{lotfian2017building}. Each turn is annotated with one of nine categorical emotion labels (anger, happiness, sadness, disgust, surprised, fear, contempt, neutral, other). The Mandarin Affective Speech (Mandarin AS) corpus contains 25,636 utterances from 68 unique speakers, with annotations according to five emotion labels (anger, elation, neutral, panic, sadness) \cite{yang2007mandarin,wu2006masc}. The IEMOCAP benchmark contains 12 hours of audio consisting of 10,039 total turns from ten unique speakers, with nine different emotion labels (anger, happiness, excitement, sadness, frustration, fear, surprise, other, neutral). All corpora consist of English speech other than Mandarin AS, which consists of Mandarin speech.
\subsection{Pre-Finetuning in Speech}
In this work, we pre-finetune base Wav2Vec2.0 (94.4M parameters), inspired the workflow used by \cite{padmakumar-etal-2022-exploring} with language models. As depicted in Figure~\ref{fig:validation_epoch_workflow}, we initialize one linear classification head for each of our pre-finetuning tasks, appropriately setting the output space. For each training step, we load an instance from a randomly sampled pre-finetuning task and map the loaded instance to the corresponding classification head. We compute a scaled loss for each task separately, as in \cite{aghajanyan-etal-2021-muppet,padmakumar-etal-2022-exploring}. The scaled loss is given as $\frac{L_i(x_i, y_i, \theta)}{ln~n(i)}$ for a model parameterized by $\theta$, where $L_i(x_i, y_i, \theta)$ is the loss for training instance $i$, and $n(i)$ is the size of the output space for the prediction task of instance $i$~\cite{aghajanyan-etal-2021-muppet}. During validation, we average the taskwise losses, which helps prevents the model from overfitting to individual tasks during pre-finetuning. We pre-finetune for up to 200 epochs with early stopping after three epochs without improvement. 

We pre-finetuned each model $j$ on one of the combinations of corpora from the power set\footnote{We use $1 + \sum_{r=1}^4 {4 \choose r} = 16$ different models downstream.} of the corpora in Section~\ref{corpora_selection}. That is, each combination is a set of corpora $C_j$ where $0 \leq |C_j| \leq 4$ and $j \in \{1..16\}$. This includes a base Wav2Vec2.0 model without any pre-finetuning as a baseline (i.e., when $|C_j| = 0$). We additionally attempted to use standard emotion recognition baselines~\cite{zhang2020multimodal} such as the ComParE 2016 automatic paralinguistic feature extractor~\cite{schuller2016interspeech} with a dense artificial neural network, but these approaches cannot surpass the performance of constant prediction. This highlights the difficulty of the few-shot version of this task.  All experiments are run using individual NVIDIA RTX A6000 GPUs. We use the HuggingFace Transformers~\cite{wolf2020transformers} and PyTorch~\cite{paszke2019pytorch} packages.

\subsection{Downstream Finetuning}
To model speaker adaptation as few-shot learning, we perform downstream finetuning on emotion recognition for each individual speaker in the Emotional Speech Dataset (ESD; \cite{zhou2021seen}). There are 10 English speakers and 10 Mandarin speakers. Each study participant was a native speaker of their respective language. Each utterance contains neutrally worded language and is categorized as either Happy, Sad, Surprised, Angry, or Neutral. We perform binary emotion classification for each emotion\footnote{The targets are whether or not an instance matches a specific emotion, e.g. Happy/Not Happy, Sad/Not Sad, etc.} for each speaker under seven few-shot settings, $k \in \{2,4,8, 16, 24, 32, 64\}$, where $k$ is the number of available training examples. For each few-shot setting, we randomly sample $\frac{k}{2}$ positive and $\frac{k}{2}$ negative training instances. We conduct three trials for each downstream condition. Each trial runs for up to 200 epochs, with 30 epochs of early stopping patience.
\begin{figure}
    \centering    
    \scalebox{0.8}{
    \includegraphics[width=\linewidth]{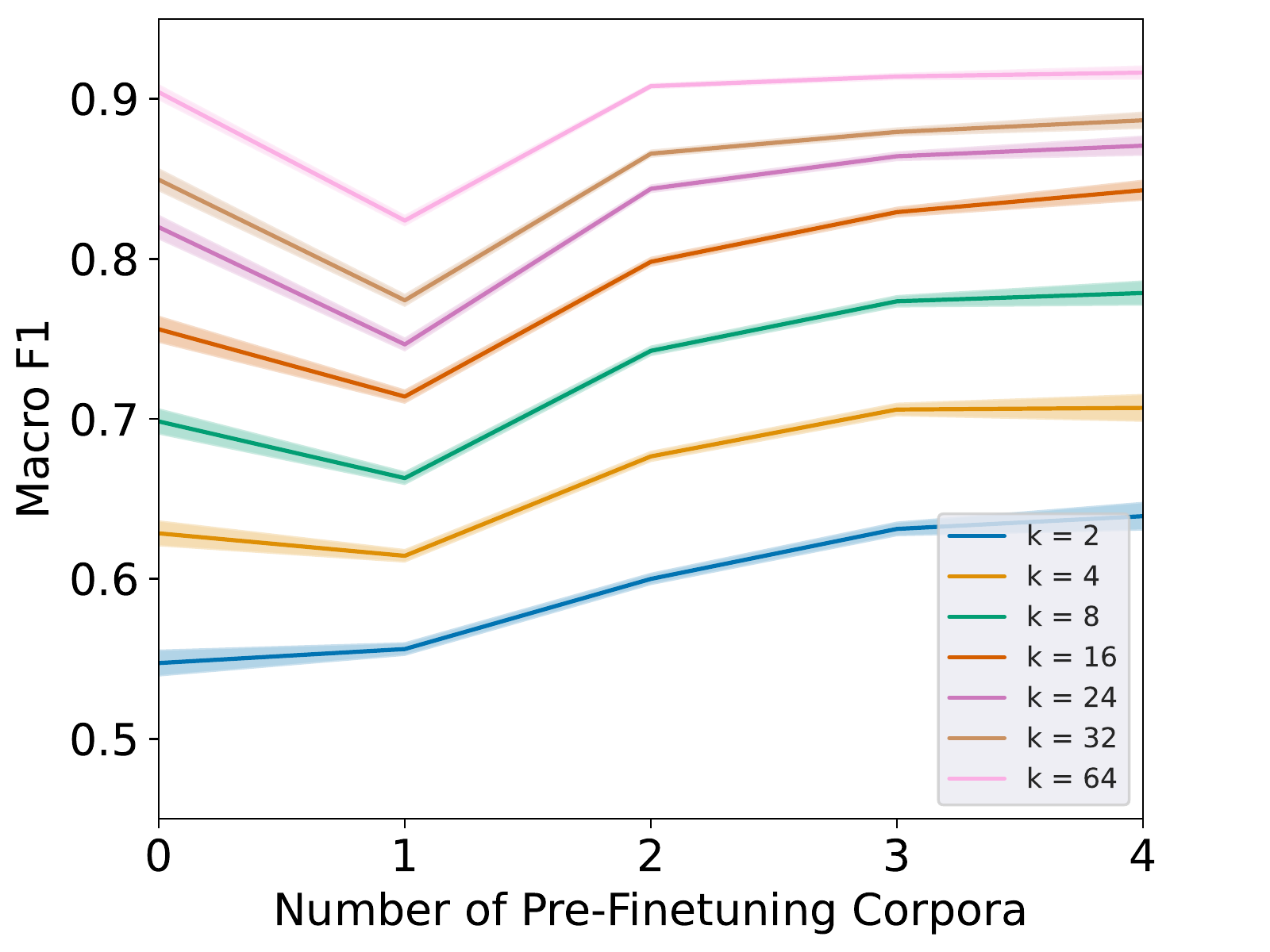}
    }
    \caption{Comparison of downstream task performance of models pre-finetuned on varying numbers of corpora. Each line depicts change in mean and standard error of F1 Macro.}
    \label{fig:prefinetuning_corpora}
\end{figure}
\section{Experimental Results}
Accounting for the 16 combinations of pre-finetuned models and all fine-tuning conditions, we evaluated the effects of pre-finetuning across 33,600 downstream model fine-tuning trials.
\subsection{Effect of Number of Pre-Finetuning Corpora}
In Figure~\ref{fig:prefinetuning_corpora} we demonstrated the effect of varying the number of corpora ($n$) used during pre-finetuning in the low-resource setting. Each line represents one few-shot condition ($k$). Each point represents the average test set F1 score of pre-finetuned models for a particular pre-finetuning corpus size for all downstream classification tasks under few-shot condition $k$. The shading around each line represents the region of standard error surrounding the mean. We observe a few patterns. In six of the seven few-shot settings, using only one pre-finetuning corpus may hurt performance compared to direct fine-tuning without a pre-finetuning step, which is consistent with the ``critical point'' for pre-finetuning utility, as discussed in \cite{aghajanyan-etal-2021-muppet}. However, in the most extreme case with $k=2$, even using just one pre-finetuning corpus still yields performance improvements over the baseline. \textit{After} $n=1$, we witness continuous improvements in average performance as $n$ increases. However we typically see the largest improvements from $n=1$ to $n=2$.  We examine possible reasons by ablating the pre-finetuning corpora.

\subsection{Ablation on Individual Corpus Contributions}
\label{sec:corpus_contributions}
We attempt to quantify how much each individual pre-finetuning corpus contributes to changes in downstream classification performance. We controlled for each speaker, each emotion, and each few-shot condition for fair comparisons. Then, under each of these controlled settings, we compute the average F1 score across all model trials for which pre-finetuning set $C_j$ includes each corpus $c$, subtracted by the average performance of the no-prefinetuning baseline. For each of these controlled settings, we calculated the change in average F1 scores for each pre-finetuning corpus compared to baseline Wav2Vec2.0. 

Figure~\ref{fig:improvement_baseline} illustrates these differentials aggregated across speakers and emotions, and stratified by each few-shot setting. We consistently see that models pre-finetuned on combinations of corpora which include MSP-PODCAST results in the most improvements over baseline performance on average, whereas IEMOCAP results in the least improvements. In the $k=32$ and $k=64$ few-shot data settings, pre-finetuning on IEMOCAP actually hurts performance on average. This is also true of MSP-IMPROV in one setting, but MSP-IMPROV is also the second most useful corpus for pre-finetuning in the two most extreme few-shot data settings. The Mandarin Affective Speech corpus consistently positively contributes to performance improvements, on average. However, the fact that certain corpora hurt performance overall may explain why there tends to be a large improvement in performance from $n=1$ to $n=2$.
\subsection{Ablation on Pre-Finetuning Corpus Inclusion}

We further attempted to isolate the contributions of individual corpora by examining the effect of including and excluding individual corpora in the extreme few-shot settings. That is, for each individual pre-finetuning corpus $c$ from the set of all corpora $C$, we compared models pre-finetuned on $C_j = C \setminus \{c\}$ and $C_j = \{c\}$. As in Section~\ref{sec:corpus_contributions}, we aggregated performances controlled for speakers, emotions, and few-shot settings, to ensure for fair direct comparisons. 

\begin{figure}
    \centering
    \scalebox{0.8}{
    \includegraphics[width=\linewidth]
    {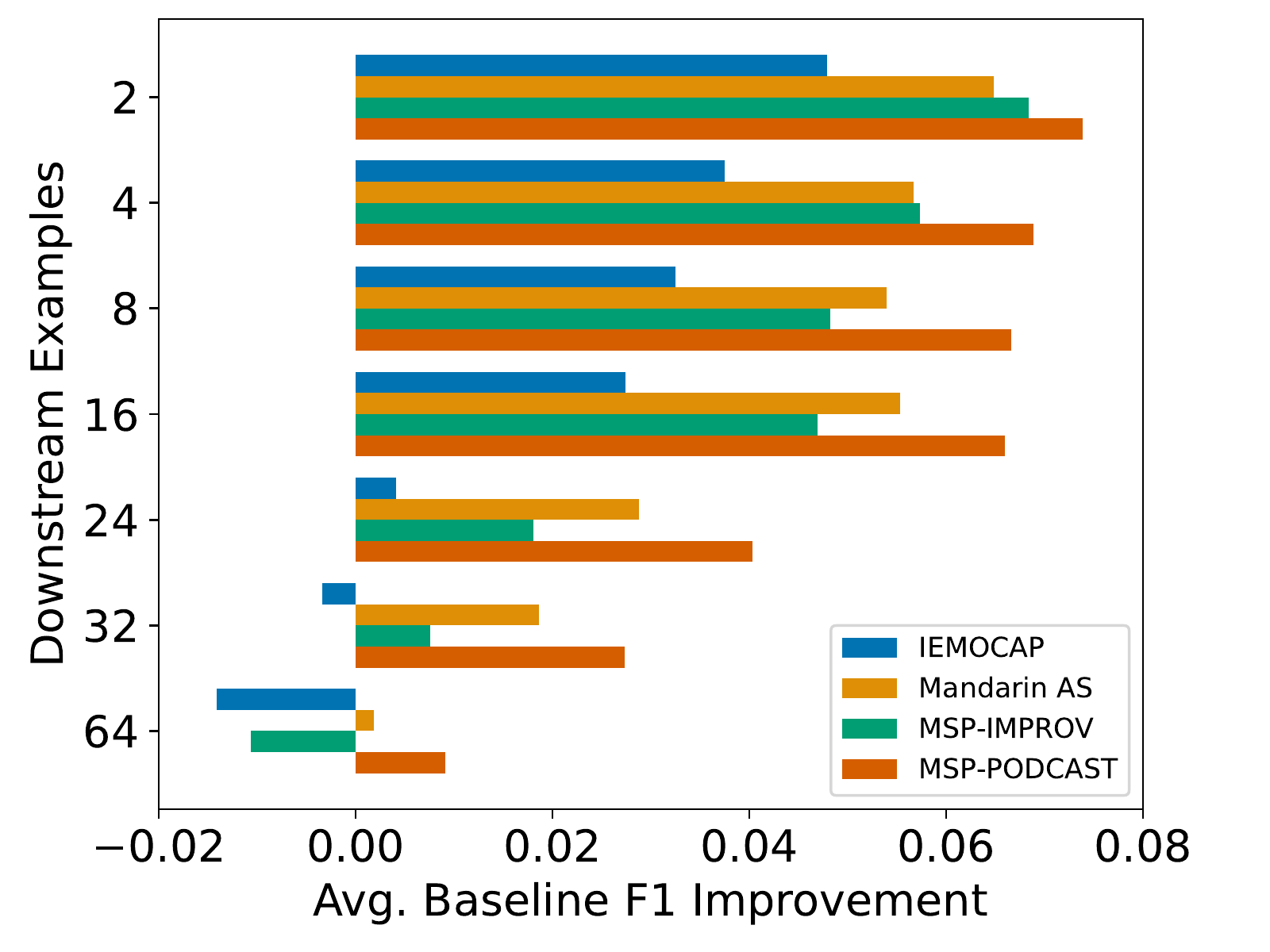}
    }
    \caption{Average difference in Macro F1 resulting from pre-finetuning on each corpus compared to the Wav2Vec2.0 baseline. Differences shown are aggregations controlling for the number of few-shot examples, each speaker, and each emotion.}
    \label{fig:improvement_baseline}
\end{figure}

In Table~\ref{tab:inclusion_exclusion_ablation}\footnote{We see the same patterns at $k \in \{16,24,32,64\}$. We omit these results due to space constraints. $F1_{in}$ for MSP-PODCAST at $k=64$ is $0.9126$ whereas $F1_{ex}$ is $0.9108$.}, we see patterns mostly consistent with the analysis of corpora contributions in Figure~\ref{fig:improvement_baseline}. $F1_{in}$ is the aggregated performance of the model trained on $C_j = \{c\}$ and $F1_{ex}$ is the performance of the model traied on $C_j= C \setminus \{c\}$. MSP-PODCAST is the only corpus which has any positive differentials when compared to pre-finetuning on all other corpora. For all few-shot settings, MSP-PODCAST has the highest differential, followed by Mandarin AS, MSP-IMPROV, and IEMOCAP.

\begin{table}[t]
\small
\begin{center}    
\caption{Average Macro F1 from pre-finetuning on each individual corpus (F1$_{in}$) compared to F1 from pre-finetuning on all but that corpus (F1$_{ex}$). F1 is aggregated controlling for the number of few-shot examples, each speaker, and each emotion.}
\label{tab:inclusion_exclusion_ablation}
\scalebox{0.75}{
\begin{tabular}{@{}lllll@{}} \toprule
$k$ & Corpus & F1$_{in}$ & F1$_{ex}$ & $\Delta$\\ \midrule
\multirow{4}{*}{\textbf{2}} & IEMOCAP & 0.5048 & 0.6232  &  -0.1184\\
& Mandarin AS & 0.5812 & 0.6445   & -0.0633   \\
& MSP-IMPROV  & 0.5233 & 0.6299  & -0.1066\\
& MSP-PODCAST & 0.6150 & 0.6272   & \textbf{-0.0122}\\
\midrule
\multirow{4}{*}{\textbf{4}} & IEMOCAP & 0.5447 & 0.7054  &  -0.1607\\
& Mandarin AS & 0.6495 & 0.7160 & -0.0665\\
& MSP-IMPROV & 0.5623 & 0.7030   & -0.1407 \\
& MSP-PODCAST & 0.7010 & 0.6990  & \textbf{~0.0020}\\
\midrule
\multirow{4}{*}{\textbf{8}} & IEMOCAP & 0.5816 & 0.7747 & -0.1931\\
& Mandarin AS & 0.7040 & 0.7862 & -0.0822\\
& MSP-IMPROV & 0.6021 & 0.7783 & -0.1762\\
& MSP-PODCAST & 0.7640 & 0.7549  & \textbf{~0.0091}\\
\bottomrule
\end{tabular}
}
\end{center}
\end{table}
\begin{figure*}
    \centering
    \scalebox{1.0}{
    \includegraphics[width=0.49\linewidth]{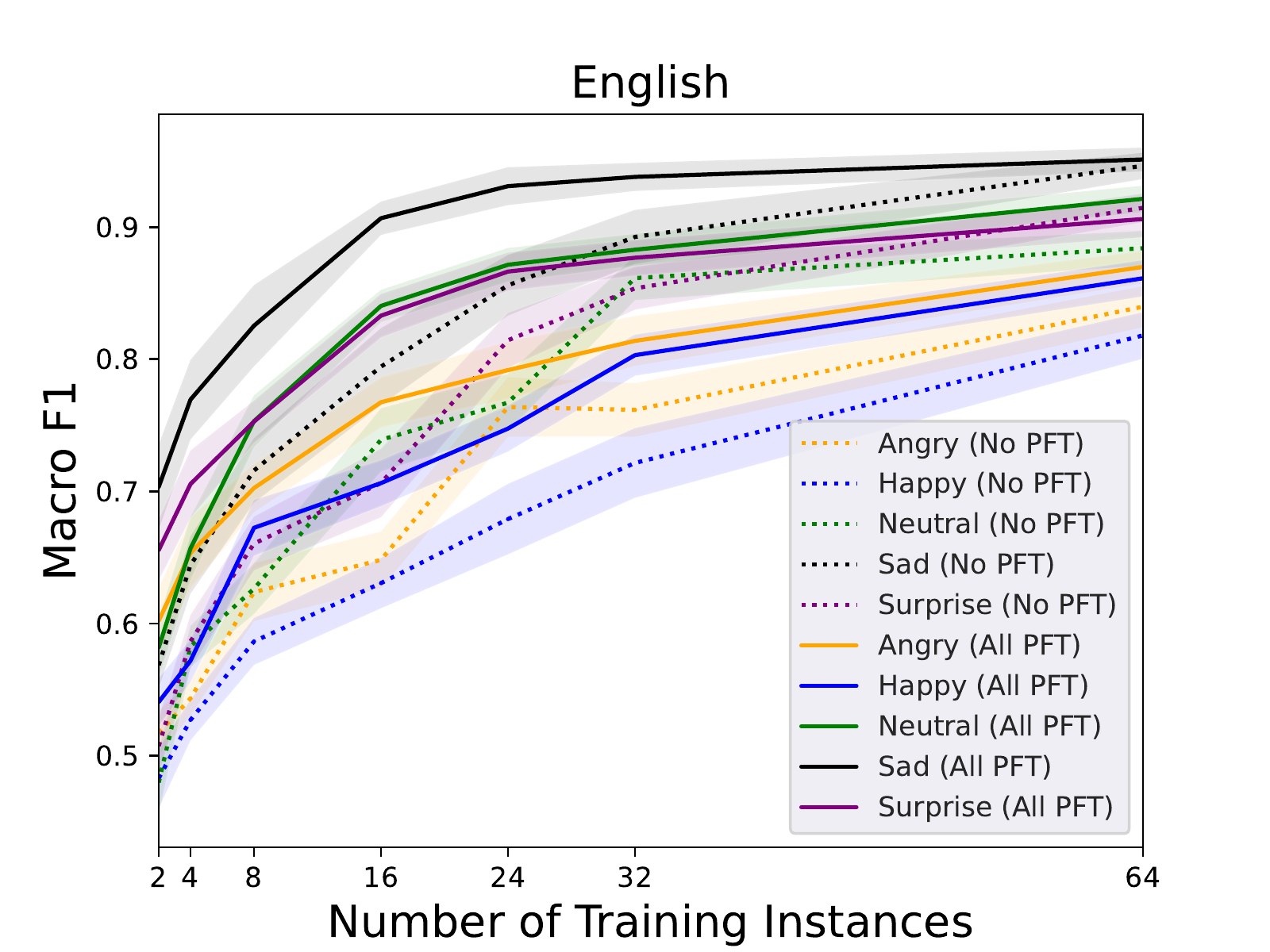}
    \includegraphics[width=0.49\linewidth]{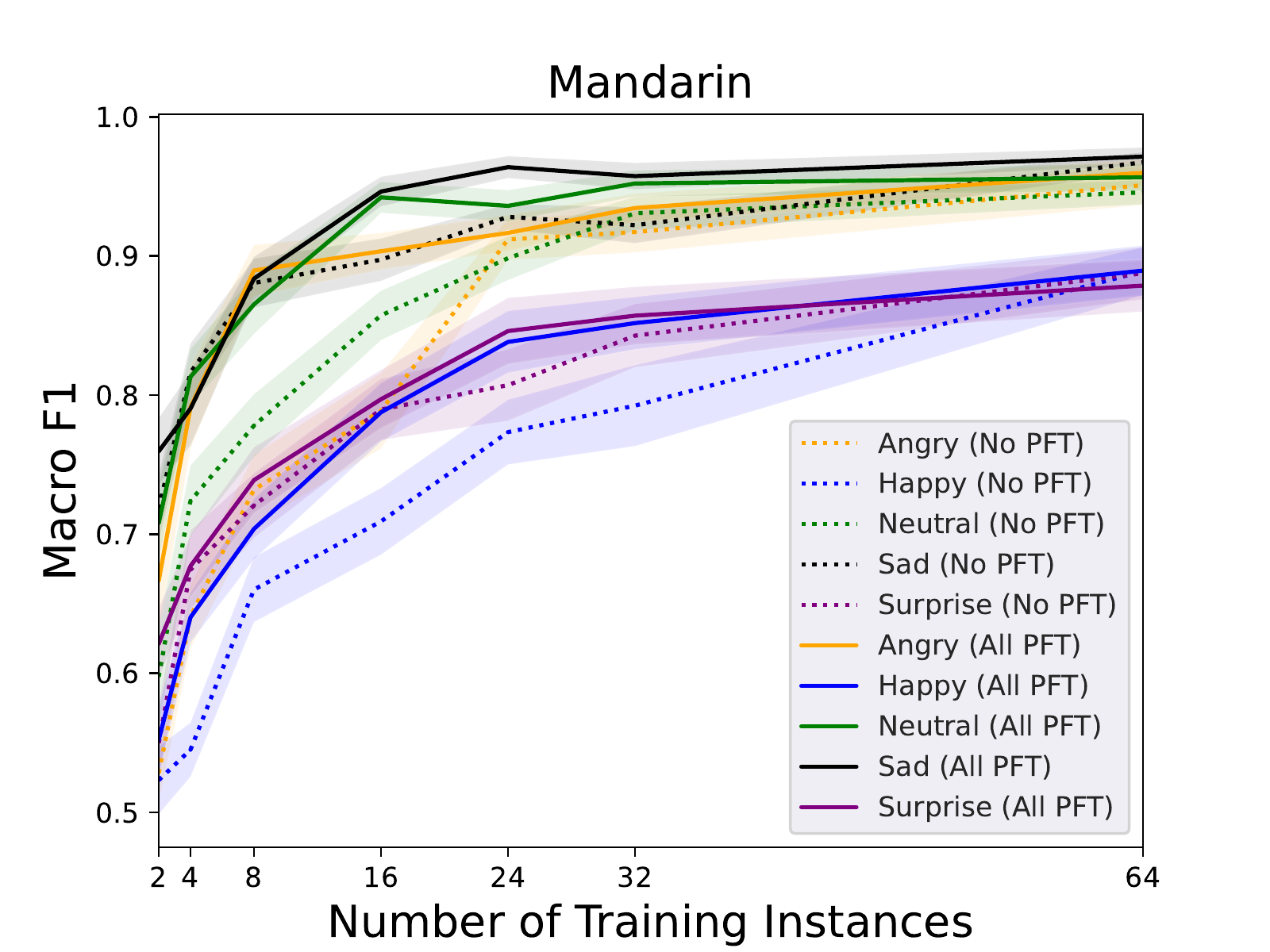}
    }
    \caption{Effect of number of training examples using during fine-tuning for the baseline model with no pre-finetuning (No PFT), and the model pre-finetuned on all four corpora (All PFT). Results are stratified by emotion. Left: classification results on native English speech. Right: classification results on native Mandarin speech.}
    \label{fig:data_settings}
\end{figure*}
\subsection{Scaling Downstream Training Data Sizes}
From Figure~\ref{fig:prefinetuning_corpora} and Figure~\ref{fig:improvement_baseline}, it is clear that the most performance improvements arise from the fewest-resource setting ($k=2$ downstream training examples), but we do see performance improvements in higher resource settings when pre-finetuning on all four corpora. We take a deeper look into these performance improvements to further understand the ceiling and limits of this pre-finetuning approach. In Figure~\ref{fig:data_settings}, we compare the performance of the full pre-finetuning setting against the no pre-finetuning baseline in all of our data settings, stratified by individual emotions and separated by speakers' native language.

For both the English and Mandarin speakers, we see that using pre-finetuning yields higher performance on average than the baseline for all few-shot data settings for all emotions except for ``Surprise.'' This may be due to the fact that whereas the other four emotions are relatively common, Surprise is not as well-represented in the pre-finetuning corpora. Surprise is only represented explicitly by MSP-PODCAST and IEMOCAP. Overall, we do see that pre-finetuning can boost classification performance substantially, but the performance gains generally taper off substantially after $k=24$.
\section{Discussion}
While conventional wisdom suggests that pre-finetuning must be performed on large-scale corpora~\cite{aghajanyan-etal-2021-muppet}, we actually show that it is possible to achieve strong results in the few-shot setting with small-scale pre-finetuning. We do observe that downstream task performance may improve further as more pre-finetuning corpora are used. In this study we only used four corpora due to licensing and computational constraints, but our findings warrant examining the upside of more corpora. A larger set may delay the onset of the diminishing returns seen in Figure~\ref{fig:data_settings}. 

We also see that the finding of a critical point in number of pre-finetuning corpora prior to witnessing performance improvements from \cite{aghajanyan-etal-2021-muppet} may hold true for speech tasks. We hypothesize that this is likely because models such as Wav2Vec2.0 adapt well to downstream fine-tuning due to their general representations learned during their masked pre-training process~\cite{baevski2020wav2vec}, while pre-finetuning on too few corpora may cause such models to lose some of their generality. We also see that MSP-PODCAST appears to contribute the most to improvements in downstream task performance. This may be due to the fact that simply is the largest corpus, both in terms of number of training instances and number of emotions covered. This is despite averaging task-specific losses during pre-finetuning.

Our experimental results reveal that the fewer the number of available downstream training examples, the more valuable pre-finetuned representations are. The $k=2$ setting is the only context in which there is not a number of pre-finetuning corpora which results in a performance decrease on average compared to Wav2Vec2.0 without pre-finetuning (Figure~\ref{fig:prefinetuning_corpora}). Moreover,  Figure~\ref{fig:prefinetuning_corpora} and Figure~\ref{fig:improvement_baseline} shows that the performance improvements over the baseline generally seem smaller the greater the number of training examples available, indicating that pre-finetuning most helps performance in the extreme low-resource settings, which reflects the initial stages of personalized speaker adaptation. This experimental setup and approach to speaker adaptation is also unconstrained by language. Our pre-finetuned models adapt to each speaker individually regardless of whether they speak English or Mandarin.

In this study, we held the model choice fixed, but our workflow is compatible with any pre-trained model. Base Wav2Vec2.0 is comparable in size to the base variants of RoBERTa and BART, so it is likely that using a larger Wav2Vec2.0 would result in improvements following similar patterns to those language models. While a benefit of pre-finetuning is that we can achieve strong performance using a small model and limited downstream data, we would likely see further performance improvements using larger models.
\section{Conclusion}
This work is the first to examine pre-finetuning on speech processing tasks. We see large performance improvements in extreme few-shot data settings, including boosting performance from near random to over $0.600$ F1 using $k=2$ training examples. We contribute an in-depth controlled analysis of several experimental factors including ablations of pre-finetuning corpora, motivating applying pre-finetuning to other speech processing tasks with more models and pre-finetuning corpora.
\section{Acknowledgements}
This work is supported by a DARPA PTG grant. We thank Ta-Chung Chi and Debasmita Bhattacharya for helpful feedback.



\bibliographystyle{IEEEtran}
\bibliography{mybib}

\end{document}